\documentclass[10pt,twocolumn,letterpaper]{article}
\usepackage{iccv}
\usepackage{times}
\usepackage{epsfig}
\usepackage{graphicx}
\usepackage{amsmath}
\usepackage{amssymb}

\usepackage{subcaption}
\usepackage{multirow}
\usepackage{bm}
\usepackage{pifont}
\newcommand{\cmark}{\ding{51}}%
\newcommand{\xmark}{\ding{55}}%

\usepackage{algorithm}
\usepackage{algorithmic}
\usepackage{color}

\def\etal{{\em et al.\/}\, }

\def\ie{\mbox{\textit{i.e.}, }}
\def\eg{\mbox{\textit{e.g.}, }}

\DeclareMathAlphabet\mathbfcal{OMS}{cmsy}{b}{n}

\def\0{{\bf 0}}
\usepackage{ dsfont }
\def\1{\mathds{1}}

\usepackage{ntheorem}

\newtheorem*{*thm}{Theorem}

\newtheorem*{*lemma}{Lemma}

\usepackage{enumitem}

\definecolor{demphcolor}{RGB}{144,144,144}
\newcommand{\demph}[1]{\textcolor{demphcolor}{#1}}

\def\x{$\times$}
\newlength\savewidth\newcommand\shline{\noalign{\global\savewidth\arrayrulewidth
  \global\arrayrulewidth 1pt}\hline\noalign{\global\arrayrulewidth\savewidth}}

\usepackage[pagebackref=true,breaklinks=true,letterpaper=true,colorlinks,bookmarks=false]{hyperref}

\iccvfinalcopy 

\def\iccvPaperID{***} 

\ificcvfinal\fi

\begin{document}

\title{ASCNet: Self-supervised Video Representation Learning \\ with Appearance-Speed Consistency}

\author{Deng Huang$^{1,3}$\thanks{ Co-first authorship. This work was done when Deng Huang was a research intern at Baidu VIS.}\qquad
Wenhao Wu$^{2*}$\qquad
Weiwen Hu$^{1}$\qquad
Xu Liu$^{1}$\vspace{.2em}\\
Dongliang He$^{2}$\quad
Zhihua Wu$^{2}$\quad
Xiangmiao Wu$^{1}$\quad
Mingkui Tan$^{1,4}$\thanks{ Corresponding author.}\quad
Errui Ding$^{2}$\vspace{.3em}\\
$^1$ South China University of Technology \quad
$^2$ Baidu Inc. \quad
$^3$ Pazhou Laboratory \\
$^4$ Key Laboratory of Big Data and Intelligent Robot, Ministry of Education \\
{\tt\small
    \{sehuangdeng,sehuww,seqdmy\}@mail.scut.edu.cn, 
    \{xmwu, mingkuitan\}@scut.edu.cn, 
}\\
{\tt\small wuwenhao17@mails.ucas.edu.cn, \{hedongliang01, wuzhihua02, dingerruis\}@baidu.com}
}

\maketitle
\ificcvfinal\thispagestyle{empty}\fi

\begin{abstract}
We study self-supervised video representation learning, which is a challenging task due to 1) lack of labels for explicit supervision; 2) unstructured and noisy visual information. Existing methods mainly use contrastive loss with video clips as the instances and learn visual representation by discriminating instances from each other, but they need a careful treatment of negative pairs by either relying on large batch sizes, memory banks, extra modalities or customized mining strategies, which inevitably includes noisy data.
In this paper, we observe that the consistency between positive samples is the key to learn robust video representation. Specifically, we propose two tasks to learn appearance and speed consistency, respectively. The appearance consistency task aims to maximize the similarity between two clips of the same video with different playback speeds. The speed consistency task aims to maximize the similarity between two clips with the same playback speed but different appearance information. We show that optimizing the two tasks jointly consistently improves the performance on downstream tasks, \eg action recognition and video retrieval.
Remarkably, for action recognition on the UCF-101 dataset, we achieve 90.8\% accuracy without using any extra modalities or negative pairs for unsupervised pretraining, which outperforms the ImageNet supervised pretrained model.
Codes and models will be available.
\end{abstract}

\begin{figure}[t]
    \centering
    \includegraphics[width=0.47\textwidth]{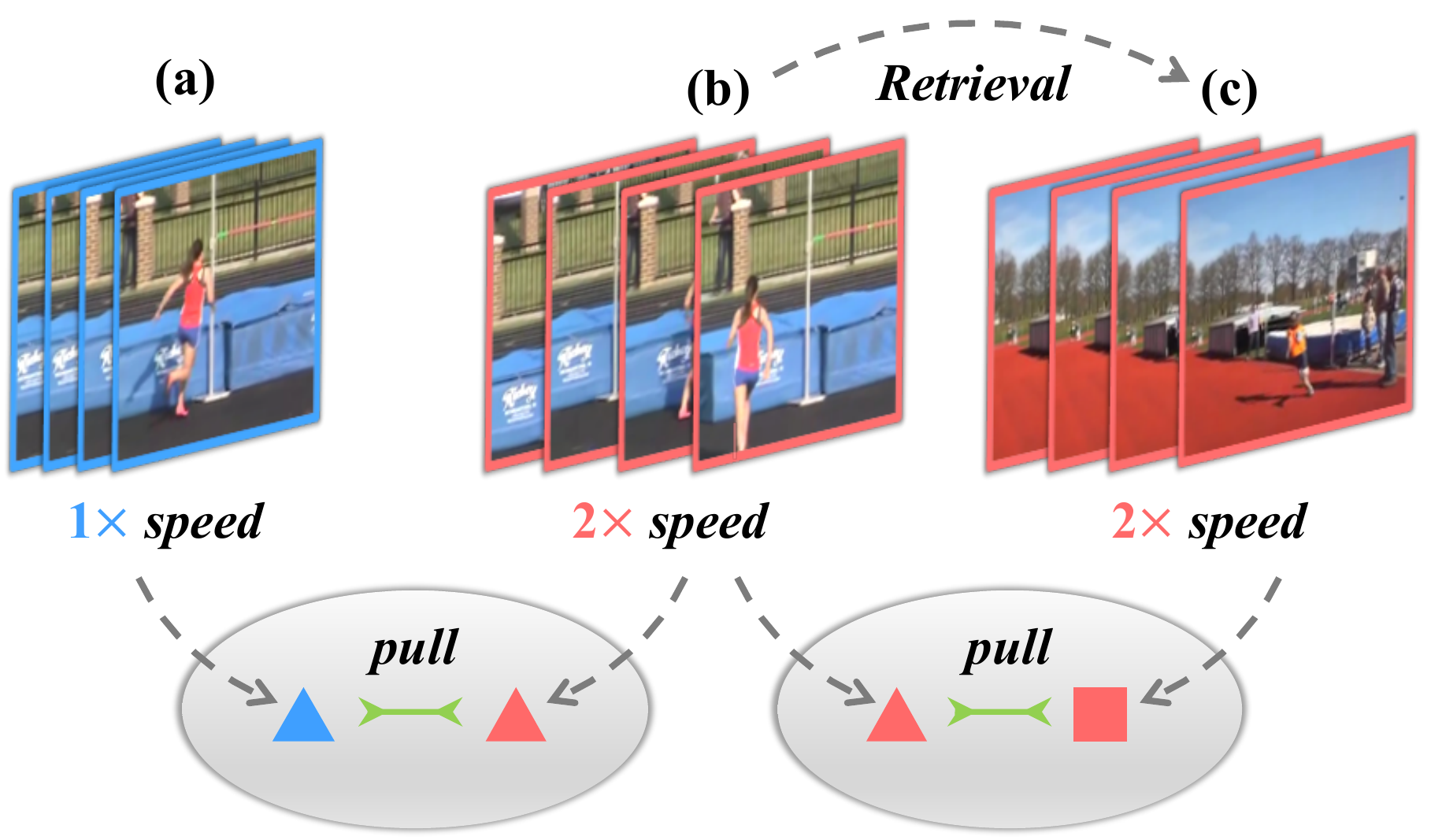}
    \caption{\textbf{An illustration of appearance-speed consistency.} Video clips (a) and (b) come from the same video, and the appearance is consistent with different playback speeds. On the other hand, by using the same frame interval, we can sample the clip (c) from different videos but with the same playback speed as clip (b). We train the model to map clips to appearance and speed embedding space while maintaining the consistency. The appearance-based retrieval strategy reduces the conflict between these two objectives.}
    \label{fig:overview}
\end{figure}

\section{Introduction}

By 2022, almost 79\% of the world's mobile data traffic will be video. With the rise of cameras on smartphones, recording videos has never been easier. Video analysis has become one of the most active research topics~\cite{MARL,MVFNet,wu2021dsanet}. However, the generation of high-quality video data requires a massive human annotation effort (\eg Kinetics-400~\cite{kay2017kinetics}, Youtube-8M~\cite{youtube8m}), which is expensive, time-consuming, and hard to scale up. In contrast, millions of unlabeled videos are freely available on the Internet, \eg YouTube. Thus, learning meaningful representations from unlabeled videos is crucial for video analysis.

Self-supervised learning makes it possible to exploit a variety of labels that come with the data for free. Instead of collecting manual labels, the proper learning objectives are set to obtain supervision from the unlabeled data themselves. These objectives, also known as \textit{pretext tasks}, roughly fall into three categories: 1) predicting specific transformations (\eg rotation angle~\cite{3DRotNet}, playback speed~\cite{speednet}, and order~\cite{ClipOrder}) of video clips; 2) generative dense prediction, \eg future frame prediction~\cite{DPC}; and 3) instance discrimination, \eg CVRL~\cite{CVRL} and Pace~\cite{pace}. Among these methods, the playback speed prediction task has attracted more attention because 1) we can easily train a model with speed labels generated automatically from the video inputs, and 2) models will focus on the moving objects to perceive the playback speed~\cite{pace}. Thus, models are encouraged to learn representative motion features.

While promising results have been achieved, existing methods suffer from two limitations.
First, some of the approaches rely on pre-computed motion information (\eg optical flow~\cite{han2020coclr,MAS}), which is computationally heavy, particularly when the dataset is scaled up.
Second, while negative samples play important roles in instance discrimination tasks, it is hard to maintain their quality and quantity.
Moreover, same-class negative samples can be harmful~\cite{cai2020negatives} to the representations used in downstream tasks.

In this work, we explore the consistency between both appearance and speed of video clips from the same and different instances and eliminate the need for negatives that are detrimental in some cases. To this end, we propose two new pretext tasks, namely,  \emph{\textbf{A}ppearance \textbf{C}onsistency \textbf{P}erception (\textbf{ACP})} and \emph{\textbf{S}peed \textbf{C}onsistency \textbf{P}erception (\textbf{SCP})}. Specifically, for the ACP task, we sample two clips from the same video with independent data augmentations and encourage the representations of the two clips to be close enough in feature space. Models cannot finish this task by learning low-level information, \ie color and rotation. Instead, models tend to learn appearance features such as background scenes and the texture of objects because these features are consistent along a video. For the SCP task, we sample two clips from two different videos with the same playback speed. Representations of these two clips are pulled closer in the feature space. Since the appearance varies from instance to instance, speed can be the crucial clue to finish this task.

Moreover, to enrich the positive samples and integrate the ACP and SCP tasks, we propose an appearance-based video retrieval strategy, which is based on the observation that appearance features in the ACP task achieve a decent accuracy (45\% top-1) in the video retrieval task. Thus, we collect the video with the same speed and similar appearance for the SCP task and make it more compatible with the ACP task. This strategy further improves the performance of downstream tasks with negligible computational cost.

To summarize, our contributions are as follows:
\begin{itemize}
    \item We propose the ACP and SCP tasks for unsupervised video representation learning. In this sense, negative samples no longer affect the quality of learned representations, making the training more robust.
    \item We propose the appearance-based feature retrieval strategy to select the more effective positive sample for speed consistency perception. In this way, we can bridge the gap between two pretext tasks.
    \item We verify the effectiveness of our method for learning meaningful video representations on two downstream tasks, namely, action recognition and retrieval, on the UCF-101~\cite{ucf101} and HMDB51~\cite{hmdb51} datasets. In all cases, we demonstrate state-of-the-art performance over other self-supervised methods, while our method is easier to apply in practice because we do not have to maintain the collection of negative samples.
\end{itemize}

\section{Related Work}

\textbf{Self-supervised image representation learning.}
Self-supervised visual representation has recently witnessed tremendous progress on images.
A common workflow is to train an encoder on one or multiple pretext tasks with unlabeled images and then use one intermediate feature layer of this model to feed a linear classifier on image classification.
The final classification accuracy qualifies how good the learned representation is.
The pretext tasks include image rotations~\cite{gidaris2018rotate}, jigsaw puzzle~\cite{norooze2016jigsaw} and colorization~\cite{zhang2016colorize}.
Recent progress of self-supervised learning is mainly based on instance discrimination, which maintains relative consistency between the representations of the anchor image and its augmented view.
The performance of contrastive learning relies on a rich set of negative samples~\cite{chen2020simclr,chen2021contrastive,he2020mocov1}.
SimCLR~\cite{chen2020simclr} uses a large batch size and picks negatives in a minibatch.
MoCo~\cite{he2020mocov1} maintains a large dictionary that covers negative features as a memory bank.
Nevertheless, same-class negatives are inevitable and harmful to the performance of contrastive learning~\cite{cai2020negatives}.
Recently, beyond contrastive learning, BYOL~\cite{grill2020byol} and SimSiam~\cite{chen2020simsiam} learn meaningful representation by only maximizing the similarity between two augmented positive samples without collapsing.
%
%

\begin{figure*}[t!]
	\centering
	\includegraphics[width=0.98\textwidth]{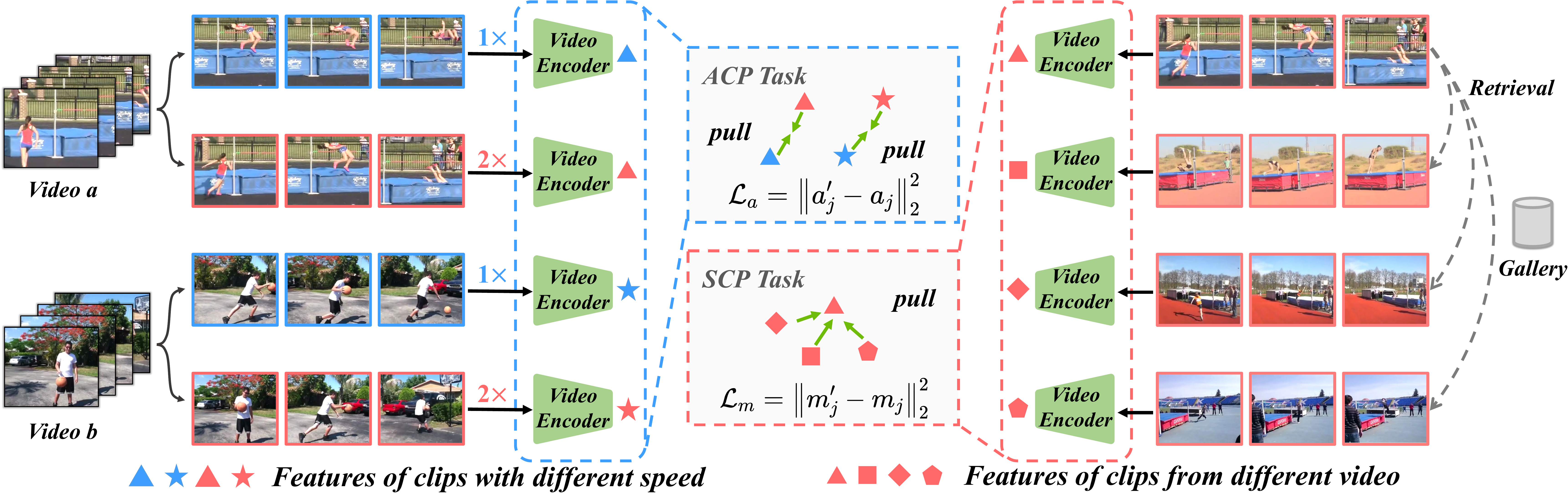}
	\caption{Illustration of the proposed framework. Given a set of video clips with different playback speed (\ie $1\times$ and $2\times$), we use a video encoder $f(\cdot; \theta)$ to map the clips into appearance and speed embedding space. For the ACP task, we pull the appearance features from the same video closer. For the SCP task, we first retrieve the same speed video with similar contents and then pull the speed features closer. All of the video encoders share the parameters.}
	\label{fig:network}
	
\end{figure*}

\textbf{Self-supervised video representation learning.} In recent years, video analysis has become a popular topic. Unlike static images, videos offer additional opportunities for learning meaningful representation by exploiting spatiotemporal information. Existing methods learn representation through various carefully designed pretext tasks. Some pretext tasks are extended from image-based representation learning, \eg rotation prediction~\cite{3DRotNet}, jigsaw~\cite{3dpuzzle}. Other approaches pay more attention to temporal information, including sorting video frames or clips~\cite{OPN}. BE \cite{wang2020BE} erases the influence of video background by mixing a static frame with the whole clip and removing it. More recently, several attempts have been made through playback speed prediction. SpeedNet~\cite{speednet} predicts whether the video clip is sped up or not, while Pace~\cite{pace} predicts the exact speed of the video clip. Instead of predicting the absolute playback speed, RSPNet~\cite{chen2020rspnet} predicts the relative speed to avoid the dependence on imprecise speed labels. 
However, some movements are too small to make a difference even under different playback speeds. Instead, we only focus on the speed similarity.

\section{Approach}

\textbf{Problem definition.} We let $\mathcal{V} = \{ v_i \}_{i=1}^N$ be a set of training videos, and we sample a clip $c_i$ from $v_i$ with playback speed $s_i$. Self-supervised video representation learning aims to learn an encoder $f(\cdot; \theta)$ to map the clip $c_i$ to consistent feature $\mathbf{x}_i$ under different video augmentations.

This task is extremely difficult because of insufficient labels and complex spatiotemporal information. First, it is difficult to construct proper supervision from unlabeled videos for models to learn both appearance and motion information. Second, it is inefficient to capture motion information from video, \eg by computing optical flow in a sequence of frames. Consequently, the learned representation may not fulfill the requirements of downstream tasks, such as action recognition and video retrieval.

\subsection{General Scheme of ASCNet}


In this paper, we observe that video playback speed not only is a good source of temporal data augmentation that does not change the spatial appearance but also provides effective supervision for video representation learning.  Thus, we propose \emph{\textbf{A}ppearance \textbf{C}onsistency \textbf{P}erception (\textbf{ACP})} task for learning appearance features, \ie predict consistent appearance features under different spatiotemporal augmentations of the same video, and the \emph{\textbf{S}peed \textbf{C}onsistency \textbf{P}erception (\textbf{SCP})} task for learning speed features, \ie predict consistent speed features for different videos with the same playback speed.

Formally, for the ACP task, different from training the model to predict whether or not two clips $c_i$ and $c_j$ are sampled from the same video, we propose to minimize the distance between the representations of clips $c_i$ and $c_j$ in the embedding space. Our intuition is that the clips sampled from the same video naturally share similar appearance contents. We also randomize the playback speed so that $s_i$ can be equal or not to $s_j$. In this way, models are encouraged to learn appearance consistency.
For the SCP task, we enforce the models to encode the playback speed information and shorten the distance between the representations of clips $c_i$ and $c_k$, which are sampled from different videos with $s_i = s_k$.
In this manner, models are encouraged to learn the information that they have in common, namely, the playback speed.


We use two individual projection heads $g_a$ and $g_m$ to map the representation from $f$ to task corresponding features $\mathbf{a}_i, \mathbf{a}_j, \mathbf{m}_i, \mathbf{m}_k$,
where $\mathbf{a}_i, \mathbf{a}_j$ are the features for the ACP task and $\mathbf{m}_i, \mathbf{m}_k$ are the features for the SCP task.
The overall objective function of our \emph{\textbf{A}ppearance-\textbf{S}peed \textbf{C}onsistency \textbf{Net}work (\textbf{ASCNet})} is formulated as follows:
\begin{equation}
    \label{eq:objective}
    \mathcal{L}(\mathcal{V}) = \gamma\mathcal{L}_m(\mathcal{V}) + (1-\gamma) \mathcal{L}_a(\mathcal{V}),
\end{equation}
where $\mathcal{L}_a$ and $\mathcal{L}_m$ denote the loss function of the ACP and SCP tasks, respectively.
$\gamma$ is a hyperparameter that controls the importance of each task.
The pretrained encoder $f(\cdot; \theta)$ and its output feature $\mathbf{x}$ will be used in downstream tasks.

\subsection{Appearance Consistency Perception}

This task aims to minimize the representation distance between two augmented clips from the same video. Given a video, we sample two clips $c_i, c_j$ with different playback speeds $s_i, s_j$, respectively. We feed the clips into the video encoder $f$ followed by a projection head $g_a$ to obtain the corresponding features $\mathbf{a}_i, \mathbf{a}_j$.
Following common practice~\cite{grill2020byol,chen2020simsiam}, we pass $\mathbf{a}_i$ to an additional predictor $h_a$ to obtain the prediction $\mathbf{a}_i'$.
We also employ a momentum target encoder, whose parameters $\xi$ is an exponential moving average (EMA) of the corresponding parameters $\theta$.
We minimize the feature distance by using $l_2$ loss as follows:

\begin{equation}
    \label{eq:appearance}
    \begin{aligned}
        \mathbf{a}_i' &= h_a(g_a(f(c_i; \theta); \theta_a); \theta_a') \\
        \mathbf{a}_j &= g_a(f(c_j; \xi); \xi_a) \\
        \mathcal{L}_a &= \left\| \mathbf{a}_i' - \mathbf{a}_j \right\|_2^2.
    \end{aligned}
\end{equation}

Since different data augmentations and playback speeds do not change the content of the clip, we expect the appearance features $\mathbf{a}_i'$ and $\mathbf{a}_j$ to be always similar.

\subsection{Speed Consistency Perception}
Temporal information is crucial for the downstream tasks, \eg action recognition. Recently, video playback speed prediction has been used as a successful pretext task for perceiving temporal information~\cite{speednet,pace}.
However, direct prediction of speed may be suboptimal for learning effective representation because the changes of some motion may be not obvious under different playback speeds. Thus, we propose the consistent speed perception task.
This task aims to minimize the distance between two clips with the same playback speed while the appearance can be different. Specifically, we sample two clips $c_i, c_k$ from two videos $v_i, v_k$ in the video set $\mathcal{V}$ with $s_i = s_k$.
Then, these two clips are processed similarly to the ACP task, except that we use projection head $g_m$ and predictor $h_m$ independent from those in the ACP task.
Finally, we define the following $l_2$ loss between the prediction and its target feature:
\begin{equation}
    \label{eq:motion}
    \begin{aligned}
        \mathbf{m}_i' &= h_m(g_m(f(c_i; \theta); \theta_m); \theta_m') \\
        \mathbf{m}_k &= g_m(f(c_k; \xi); \xi_m) \\
        \mathcal{L}_m &= \left\| \mathbf{m}_i' - \mathbf{m}_k \right\|_2^2.
    \end{aligned}
\end{equation}
However, for two dissimilar videos, the optimization of $\theta$ can be difficult, and it takes more time for the model to converge. Thus, we propose an appearance-based feature retrieval framework to collect similar videos in feature space.

\subsubsection{Appearance-based Feature Retrieval}


The \textbf{instance sampling strategy} affects the performance of both the SCP and ACP tasks. The clips used in the SCP task can be sampled from the \textbf{same instance} or \textbf{different instances}. However, when using the former, the SCP task may fall back into the ACP task. Since some movements have their corresponding speed, \ie running and jogging, the latter may lead to conflict between the ACP and SCP tasks. Thus, to reduce the conflict, we propose an appearance-based feature retrieval strategy as follows.

Given an anchor video $v_i$, we collect a candidate set of videos (gallery) $\mathcal{C} = \{ v_1, v_2, ..., v_t \}_{t=1}^T$ from $\mathcal{V} \setminus v_i$. We sample a clip from each video to obtain the anchor feature $\mathbf{a}_i$ and candidate features $\{ \mathbf{a}_1, \mathbf{a}_2, ..., \mathbf{a}_t \}_{t=1}^T$ by using the same process as the ACP task. A simple dot product function $d(\cdot, \cdot)$ is used to measure the similarity between the anchor and candidates. Then, we sort the videos by their similarity score and select $v_k$ from the most similar candidates. In practice, we use the memory bank~\cite{he2020mocov1} to reduce the computation cost. The video pair $v_i, v_k$ can be used in the SCP task with the benefit of strong spatial augmentation while not breaking the appearance consistency.

\begin{algorithm}[ht]
\caption{Training method of ASCNet.}
\label{algo:training}
\begin{algorithmic}[1]
    \REQUIRE Video set $\mathcal{V} = \{ v_i \}_{i=1}^N$,
    the encoder $f$ with paramaters $\theta$ or $\xi$,
    the projection heads $g_a$ and $g_m$ with parameters $\theta_a$, $\xi_a$, $\theta_m$ and $\xi_m$,
    the predictors $h_a$ and $h_m$ with parameters $\theta_a'$ and $\theta_m'$,
    the hyperparameter $\gamma$.
    \STATE Randomly initialize parameters $\theta, \theta_a, \theta_m, \theta_a',  \theta_m'$.
    \STATE Initialize parameters $\xi\gets\theta, \xi_a\gets\theta_a, \xi_m\gets\theta_m$.
    \WHILE{not convergent}
        \STATE Randomly sample a video $v$ from $\mathcal{V}$.
        \STATE Sample two clips $c_i, c_j$ from $v$.
        \STATE Extract features $\mathbf{x}_i{=}f(c_i, \theta)$, $\mathbf{x}_j{=}f(c_j, \xi)$.
        \STATE // \emph{\textbf{Learn appearance features with ACP task}}
        \STATE Obtain $\mathbf{a}_i{=}g_a(\mathbf{x}_i, \theta_a)$, $\mathbf{a}_j{=}g_a(\mathbf{x}_j, \xi_a)$.
        \STATE Obtain $\mathbf{a}_i'=h_a(\mathbf{a}_i, \theta_a')$.
        \STATE Compute $\mathcal{L}_a=\left\| \mathbf{a}_i' - \mathbf{a}_j \right\|_2^2$ in Eqn.~(\ref{eq:appearance}).
        \STATE // \emph{\textbf{Conduct appearance-based feature retrieval}}
        \STATE Construct $\mathcal{C}{=}\{\mathbf{a}_t \}_{t=1}^{N-1}$ using $g_a(\cdot, \theta_a)$ from $\mathcal{V} \setminus v$. \\
        // \emph{\textbf{Obtain $\mathcal{C}$ efficiently from memory bank}}
        \STATE Select the video $\hat{v}$ corresponding to features $\mathbf{a} \in \mathcal{A}$ with the highest dot product similarity with $\mathbf{a}_j$.
        \STATE Sample one clip $c_k$ from $\hat{v}$.
        \STATE Extract features $\mathbf{x}_k{=}f(c_k, \xi)$.
        \STATE // \emph{\textbf{Learn speed features with SCP task}}
        \STATE Obtain $\mathbf{m}_i{=}g_m(\mathbf{x}_i, \theta_m)$, $\mathbf{m}_k{=}g_m(\mathbf{x}_k, \xi_m)$.
        \STATE Obtain $\mathbf{m}_i'{=}h_m(\mathbf{x}_i, \theta_m')$.
        \STATE Compute $\mathcal{L}_m{=} \left\| \mathbf{m}_i' - \mathbf{m}_k \right\|_2^2$ in Eqn.~(\ref{eq:motion}).
        \STATE Compute $\mathcal{L}= \gamma\mathcal{L}_m + (1-\gamma)\mathcal{L}_a$ in Eqn~(\ref{eq:objective}).
        \STATE Update parameters $\theta, \theta_a, \theta_m, \theta_a',  \theta_m'$ via SGD.
        \STATE Compute exponential moving average $\xi, \xi_a, \xi_m$.
    \ENDWHILE
\end{algorithmic}
\end{algorithm}



\section{Experiments}

\subsection{Datasets}
We consider four video datasets, including Mini-Kinetics-200~\cite{s3dg}, Kinetics-400~\cite{kay2017kinetics}, UCF-101~\cite{ucf101}, and HMDB-51~\cite{hmdb51}.
For the self-supervised pretraining, we use the training split of the Kinetics-400 dataset by discarding all of the labels.
The Kinetics-400 dataset contains 400 human action categories and provides 240k training video clips and 20k validation video clips. 
The Mini-Kinetics-200 dataset consists of 200 categories with the most training examples and is a subset of the Kinetics-400 dataset.
Since the full Kinetics-400 is quite large, we use Mini-Kinetics-200 to reduce training costs in ablation experiments.

The learned network backbones are evaluated via two downstream tasks: action recognition and nearest neighbor retrieval. For the downstream tasks, UCF-101~\cite{ucf101} and HMDB-51~\cite{hmdb51} are used to demonstrate the effectiveness of our method. UCF-101~\cite{ucf101} contains 13k videos spanning over 101 human actions. HMDB-51~\cite{hmdb51} contains approximately 7k videos belonging to 51 action classes categories.
Both UCF-101~\cite{ucf101} and HMDB-51~\cite{hmdb51} come with three predefined training and testing splits. Following prior work~\cite{ClipOrder,speednet,pace}, we use a training/testing split of 1 for downstream task evaluation.
Both datasets exhibit challenges including intraclass variance of actions, cluttered backgrounds, and complex camera motions.
Performing action recognition and retrieval on these datasets requires learning rich spatiotemporal representation.

\subsection{Implementation Details}
\begin{table*}
		\begin{subtable}[th]{0.35\textwidth}
		\centering
			\begin{tabular}{c|c}
			\shline
			Pretraining Settings  & Accuracy   \\ \hline
				w/o pretraining  & 42.40\%   \\ 
				ACP Only  & 64.76\%  \\
				SCP Only  & 43.40\%  \\ \hline
				ASCNet  & 70.71\% \\
				\shline
			\end{tabular}   
			\caption{\textbf{Study on the effectiveness of ASCNet}. 
			\emph{w/o pretraining} denotes training from scratch (random initialization) variants.
            Backbone: 3D R18.
			}
			\label{tab:ablation:perf}
		\end{subtable}	
	    \hspace{2mm}
		\begin{subtable}[th]{0.40\textwidth}
		\centering
			\begin{tabular}{c|cc}
			\shline
			Method & Configuration  & Accuracy    \\ \hline
			ACP + SP & - & 68.93\% \\
			ACP + SCP & Same instance & 69.20\% \\
			ACP + SCP & Different instances & 69.55\% \\ \hline
			ACP + SCP & Similar instances & 70.71\%  \\
			\shline
			\end{tabular}
			\caption{\textbf{Comparison with different configurations of the SCP task}. SP denotes the speed prediction task~\cite{speednet} for each video clip. Backbone: 3D R18.}
			\label{tab:ablation:ab}
		\end{subtable}	    
		\hspace{2mm}
		\begin{subtable}[th]{0.21\textwidth}
		\centering
			\begin{tabular}{c|c}
			\shline
			\{$S_1$, $S_2$\}  & Accuracy     \\ \hline
			\{\x1, \x2\}   & 70.71\%  \\ \hline
		    \{\x1, \x1\}  & 64.52\% \\
			\{\x1, \x4\}   & 70.50\%  \\
			\{\x4, \x8\}   & 72.16\%  \\
			\shline
			\end{tabular}
			\caption{\textbf{Study on different playback speeds set}. Backbone: 3D R18. }
			\label{tab:ablation:speed_ratio}
		\end{subtable}		
		\\[7pt]
		\hspace{2mm}
		\begin{subtable}[th]{0.23\textwidth}
		\centering
			\begin{tabular}{c|c}
			\shline
			Batch size  & Accuracy     \\ \hline
		    1024  & 72.20\% \\
			512   & 72.16\%  \\
			256   & 72.13\%  \\
			\shline
			\end{tabular}
			\caption{\textbf{Study on different batch sizes} used in pretraining (200 epochs). Backbone: 3D R18.}
			\label{tab:ablation:batchsize}
		\end{subtable}	
		\hspace{2mm}
		\begin{subtable}[th]{0.32\textwidth}
		\centering
			\begin{tabular}{l|c}
			\shline
			Augmentation  & Accuracy     \\ \hline
		    Color jittering  & 62.43\% \\
			+ Gaussian blurring  & 64.55\%  \\
			+ Random grayscale & 67.22\%  \\
			+ Solarization & 72.16\% \\
			\shline
			\end{tabular}
			\caption{\textbf{Ablation on data transformations for pretraining}. Backbone: 3D R18.}
			\label{tab:ablation:aug}
		\end{subtable}
		\hspace{2mm}
		\begin{subtable}[th]{0.41\textwidth}
		\centering
			\begin{tabular}{l|ccc}
			\shline
			Backbone   & Params & Random & Ours     \\ \hline
			3D ResNet-18  & 33.6M & 42.40\% &  72.16\% \\
			R(2+1)D  & 14.4M & 56.00\% & 75.95\%  \\
			S3D-G	& 9.6M & 45.31\% & 75.04\% \\ \shline
			\end{tabular}
			\caption{\textbf{Evaluation of ASC on UCF-101 using different video encoders}. We sample 16 frames with 112\x112 spatial size for pretraining and fine-tuning.}
			\label{tab:ablation:backbone}
		\end{subtable}		
		\caption{Ablation studies. All models are pretrained with 200 epochs on the \textbf{Mini-Kinetics-200}, except for w/o pretraining setting and evaluation on \textbf{UCF-101} action recognition by fine-tuning the entire network. Top-1 accuracy is reported.
		}
		\label{tab:ablations}
	\end{table*}

\textbf{Backbone networks.}
To study the effectiveness and generalization ability of our method in detail, we choose three different backbone networks as the video encoder, which have been widely used in recent video self-supervised learning methods, \ie 3D ResNet~\cite{3dresnet}, R(2+1)D~\cite{r2+1d}, and S3D-G~\cite{s3dg}.
3D ResNet~\cite{3dresnet} is a natural extension of the ResNet architecture~\cite{resnet} for directly tackling 3D volumetric video data by extending 2D convolutional kernels to the 3D counterparts.
R(2+1)D~\cite{r2+1d} is proposed to decompose the full 3D convolution into the 2D spatial convolution followed by the 1D temporal convolution.
Moreover, following previous work~\cite{speednet,pace}, we use the state-of-the-art backbone S3D-G~\cite{s3dg} to further exploit the potential of the proposed approach.

\textbf{Self-supervised pretraining stage.}
Following prior work~\cite{pace,speednet,chen2020rspnet}, we sample 16 consecutive frames with 112 \x 112 spatial size for each clip unless specified otherwise. Video clips are augmented using random cropping with resizing, random color jittering, random Gaussian blurring, and random grayscale and solarization. We utilize LARS as the optimizer with a momentum of 0.9 and weight decay of 1e-6 for training without dropout operation.
We set the base learning rate to 0.3, scaled linearly with the batch size \emph{b}; \ie the learning rate is set to 0.3\x \emph{b}/128.
The pretraining process is carried out for 200 epochs by default.
The possible playback speed \emph{s} for the clips in this paper is set to \{1\x, 2\x, 4\x, 8\x\}, \ie, sampling frames consecutively or setting the sampling interval to \{2,4,8\} frames.
We use only the raw unfiltered RGB video frames as the input and do not make use of optical flow or other auxiliary signals during training.
Moreover, we instantiate all projection heads as a fully connected layer with 256 output dimensions.
We apply L2 normalization for all features.
After pretraining, we drop the projection heads and use the features for downstream tasks. When jointly optimizing $\mathcal{L}_a$ and $\mathcal{L}_m$, we empirically set the parameters $\gamma$ as 0.5 for loss balance.

\textbf{Supervised fine-tuning stage.}
Regarding the action recognition task, during the fine-tuning stage, the learning rate was decayed by a factor of 0.01 over the course of training using cosine annealing.
The weights of convolutional layers are retained from the learned representation model, while the weights of the newly appended fully connected layers are randomly initialized.
The whole network is then trained with cross-entropy loss.

\textbf{Evaluation.}
During inference, following the common evaluation protocol, we sample 10 clips uniformly from each video in the test sets of UCF-101 and HMDB-51. For each clip, we only simply apply the center-crop instead of the ten-crop. Finally, we average the softmax probabilities of all clips as the final prediction.

\subsection{Ablation Studies}

\label{sec:ab}
As shown in Table~\ref{tab:ablations}, we provide ablation studies on the effectiveness of the different aspects of our method by self-supervised learning on Mini-Kinetics-200. The representations are evaluated on UCF-101 with end-to-end fine-tuning. The analysis was carried out as follows.

\textbf{Effectiveness of ASC.}
In this paper, we propose two tasks to learn effective video representation, namely, \emph{Appearance Consistency Perception} (\textbf{ACP}) and \emph{Speed Consistency Perception} (\textbf{SCP}).
 To verify the effectiveness of our method, we pretrain these models with 3D ResNet-18.
As shown in Table~\ref{tab:ablation:perf}, compared with training from scratch, pretraining with only the ACP task can significantly improve the action recognition performance (64.76\% \emph{vs.} 42.40\%) on the UCF-101 dataset, while consistent speed perception further improves the performance from 64.76\% to 70.71\%, indicating the effectiveness of cooperative work of these two tasks.
In the following ablation experiments, unless stated otherwise, we apply 3D ResNet-18 (3D R18) as the backbone.

\textbf{Ablation on SCP tasks.}
Here, we instantiate some variants of our method by using different speed perception tasks~\cite{speednet}.
\textbf{SP} denotes speed prediction for each individual clip.
Table~\ref{tab:ablation:ab} shows that the speed consistency perception task improves the performance compared with directly predicting the playback speed of each clip (70.71\% \emph{vs.} 68.93\%).
Then, we investigate the instance sampling strategy for the SCP task. The video clips used in the SCP task can be sampled from the \textbf{same instance} or \textbf{different instances}. \textbf{Similar instances} denotes using the appearance-based feature retrieval strategy.
These results demonstrate that the appearance-based feature retrieval strategy can benefit the speed consistency perception task while not breaking the appearance consistency.

\textbf{Different playback speed.}
We denote the $s_i$, $s_j$ in Algorithm~\ref{algo:training} as \{$S_1$, $S_2$\}.
As shown in Table~\ref{tab:ablation:speed_ratio}, we compare the performance of different playback speed sets \{$S_1$, $S_2$\} for our method.
In particular, when the speed set is \{\x1, \x1\}, our ASC loses the speed perception and degenerates to pay more attention to learning appearance information. As expected, the performance decreases from 70.71\% of \{\x1, \x2\} to 64.52\%, which is similar to the 64.76\% of \emph{ACP Only} in Table~\ref{tab:ablation:perf}.
Then, when the playback speed $S_1$ is set to \x1, we observe that the changes in $S_2$ = \x2, \x4 appear to have little impact on performance.
Interestingly, for \{\x4, \x8\}, larger sampling intervals encourage the model to explore longer motion information, boosting the learned representation (70.71\% \emph{vs.} 72.16\%).
Thus, we adopt it in the following experiments.

\textbf{Impact of batch size.}
The ablation study of different batch sizes is shown in Table~\ref{tab:ablation:batchsize}.
When the batch size changes, we use the same linear scaling rule for all batch sizes studied.
Our method works reasonably well over the wide range of batch sizes without using negative pairs.
Our experimental results show that a batch size of 256 already achieves high performance.
The performance remains stable over a range of batch sizes from 256 to 1024, and the differences are at the level of random variations.

\begin{table*}[th]
\centering
\scalebox{1}{
	\centering
		\begin{tabular}{ll|lcccc|cc} \shline
		Method & Date & Dataset (duration) & Backbone & Frames & Res. & Single-Mod  & UCF  & HMDB  \\ \hline
		Shuffle\&Learn~\cite{shuffle_and_learn}  & 2016\hspace{1pt}       & UCF (1d)      &  CaffeNet & -   & 224   & \cmark  & 50.2 & 18.1  \\ 
		OPN~\cite{OPN}        & 2017\hspace{1pt}       & UCF (1d)      &  CaffeNet   & -  & 224   & \cmark & 56.3 & 22.1  \\ 	
		CMC~\cite{cmc}  & 2019\hspace{1pt}       & UCF (1d)      &  CaffeNet  & -  & 224    & \cmark  & 59.1 & 26.7  \\ 	
		MAS~\cite{MAS}    & 2019\hspace{1pt}       & UCF (1d)      & C3D  &  16  & 112  & \xmark & 58.8 & 32.6  \\ 
		VCP~\cite{vcp}  & 2020\hspace{1pt}       & UCF (1d)      & C3D  & 16    & 112   & \cmark  & 68.5 & 32.5  \\ 
		
		ClipOrder~\cite{ClipOrder}    & 2019\hspace{1pt}       & UCF (1d)   &   R(2+1)D & 16   & $112$ & \cmark & 72.4 & 30.9  \\ 
		PRP~\cite{PRP}  & 2020\hspace{1pt}       & UCF (1d)      & R(2+1)D   & 16 &  112    & \cmark  & 72.1 & 35.0  \\ 
		PSP~\cite{psp}  & 2020\hspace{1pt}       & UCF (1d)      & R(2+1)D  & 16 & 112   & \cmark  & 74.8 & 36.8  \\ 			
		MAS~\cite{MAS}    & 2019\hspace{1pt}       & K400 (28d)      & C3D   & 16  & 112  & \xmark & 61.2 & 33.4  \\ 
		3D-RotNet~\cite{3DRotNet} & 2018\hspace{1pt}      & K400 (28d)    &  3D R18  & 16    &  $112$  & \cmark  & 62.9 & 33.7  \\ 
		ST-Puzzle~\cite{3dpuzzle}  & 2019\hspace{1pt}       & K400 (28d)    &  3D R18  & 48   & $224$    & \cmark & 65.8 & 33.7 \\ 
		DPC~\cite{DPC}        & 2019 \hspace{1pt}     & K400 (28d)    &  3D R18  & 64  & $128$  & \cmark & 68.2 & 34.5  \\ 
		
		CBT~\cite{CBT}        & 2019\hspace{1pt}      & K600+ (273d)  &   S3D-G   &  - & $112$   & \cmark & 79.5 & 44.6  \\
		SpeedNet~\cite{speednet} & 2020 \hspace{1pt}   & K400 (28d)    &   S3D-G & 64  & $224$ &  \cmark & 81.1 & 48.8  \\		
		Pace~\cite{pace} & 2020 \hspace{1pt}   & K400 (28d)    & S3D-G   & 64 &  224  & \cmark & 87.1 & 52.6  \\
		CoCLR-RGB~\cite{han2020coclr}  & 2020 \hspace{1pt} & K400 (28d)    &   S3D-G  & 32 & $128$    & \xmark   & {87.9} &  {54.6}   \\	
		RSPNet~\cite{chen2020rspnet} & 2021 \hspace{1pt} & K400 (28d)    & S3D-G   & 64  & 224 & \cmark  & {89.9} &  {59.6}   \\			
		
		\hline
		Ours &  \hspace{1pt} & K400 (28d)    &  3D R18  & 16  & $112$   & \cmark    & 80.5 &  52.3   \\
		Ours &  \hspace{1pt} & K400 (28d)    &   S3D-G & 64  & $224$   & \cmark   & \textbf{90.8} &  \textbf{60.5}   \\ \hline
	 \demph{Fully Supervised~\cite{3dresnet}} & \hspace{1pt}  &  \demph{K400 (28d)}    & \demph{3D R18}  & \demph{16}  & \demph{$112$}     & \demph{\cmark}  & \demph{84.4} & \demph{56.4}  \\
	 \demph{Fully Supervised~\cite{s3dg}} & \hspace{1pt}  &  \demph{ImageNet}    & \demph{S3D-G}  & \demph{64}  & \demph{$224$}     & \demph{\cmark}  & \demph{86.6} & \demph{57.7}  \\	
	 \demph{Fully Supervised~\cite{s3dg}} & \hspace{1pt}  &  \demph{K400 (28d)}    & \demph{S3D-G}  & \demph{64}  & \demph{$224$}     & \demph{\cmark}  & \demph{96.8} & \demph{75.9}  \\			
		\shline
		\end{tabular}
		}
	\caption{Comparison with state-of-the-art self-supervised learning methods on the UCF-101 and HMDB-51  datasets. The dataset parentheses show the total video duration (\textbf{d} for day, \textbf{y} for year). Single-Mod denotes the Single RGB Modality. K400 represents  Kinetics-400.
	}
	\label{table:sota-cls}
\end{table*}

\begin{table}[th]
\centering
\begin{tabular}{ccccc}
\shline

Arch. & Res. & \#Frames & Crop Type & Top-1 \\ \hline
\multirow{4}{*}{S3D-G} & 224 & 64 & Center-crop & 90.77\% \\
   & 224 & 64 & Three-crop & 90.88\% \\
   & 128 & 32 & Ten-crop & 87.31\% \\ \hline
\multirow{3}{*}{3D R18} & 112 & 16 & Center-crop & 80.52\% \\
   & 112 & 16 & Three-crop & 80.73\% \\   
   & 128 & 16 & Three-crop & 80.99\% \\   
\shline
\end{tabular}

\caption{Performance of different evaluation protocols. The models are pretrained with 200 epochs on Kinetics-400.}
\label{t:test}
\end{table}

\begin{table}[t]
\centering
\begin{tabular}{ccccc}
\shline
Epochs & 100 & 200 & 300 & 400 \\ \hline
Top-1 (\%) &  76.34 & 80.52 & 81.31 & \textbf{81.50}  \\ 
\shline
\end{tabular}

\caption{Performance of different pretraining epochs. A 3D ResNet-18 backbone with ASC pretraining is used. }
\label{t:epoch}
\end{table}

\textbf{Augmentation.}
The accuracies for applying the following data augmentations during pretraining one-by-one are shown in Table~\ref{tab:ablation:aug}.
With only color jittering, our ASC yields 62.43\% accuracy.
Then, we randomly blur the frames using a Gaussian distribution, boosting the accuracy by 2.1\%.
Random grayscale is the augmentation approach that converts the frames to grayscale with probability $p$ (default 0.2 in this paper). Equipped with random grayscale, the accuracy of ASC is improved from 64.55\% to 67.22\%.
Finally, we solarize RGB/grayscale video frames by inverting all of the pixel values above a threshold, further improving the performance of our ASC to 72.16\%.
Overall, by stacking these augmentations, we have steadily improved the learned representation model from 62.42\% to 72.16\%. Thus, all these data transformations are used in our experiments.

\textbf{Different backbone.}
Since it is a general framework, ASC can be widely applied to existing video backbones with consistent gains in performance. In Table~\ref{tab:ablation:backbone}, we compare various instantiations of our framework and show that our method is simple yet effective.
We observe a consistent improvement of between 20\% and 30\% on UCF-101 with our ASCNet on three video decoders, \ie 3D ResNet-18~\cite{3dresnet}, R(2+1)D~\cite{r2+1d}, and S3D-G~\cite{s3dg}.

\subsection{Evaluation on the Action Recognition Task}

\begin{table*}[t]
	\centering
	\begin{tabular}{ccccccc}
		\shline
		\multirow{2}[0]{*}{Method} & \multirow{2}[0]{*}{Architecture} &
		\multicolumn{5}{c}{Top-$k$} \\
		\cline{3-7}
		& & $k=1$    & $k=5$    & $k=10$   & $k=20$   & $k=50$     \\ 
		\hline
		OPN~\cite{OPN}      & CaffeNet             & 19.9 & 28.7 & 34.0 & 40.6 & 51.6   \\
		Buchler~\etal\cite{buchler2018improving}  & CaffeNet          & 25.7 & 36.2 & 42.2 & 49.2 & 59.5   \\
		ClipOrder~\cite{ClipOrder}    & 3D R18              & 14.1 & 30.3 & 40.0 & 51.1 & 66.5   \\
		SpeedNet~\cite{speednet} & S3D-G             & 13.0 & 28.1 & 37.5 & 49.5 & 65.0   \\ \hline
		\multirow{2}{*}{VCP~\cite{vcp}}    
		& 3D R18               & 18.6           & 33.6         & 42.5         & 53.5           & 68.1   \\
		& R(2+1)D           & 19.9  & 33.7  & 42.0 & 50.5  & 64.4   \\ \hline
		\multirow{2}{*}{Pace~\cite{pace}}    
		& 3D R18          & 23.8  & 38.1 & 46.4 & 56.6  & 69.8   \\
		& C3D               & 31.9  & 49.7  & 59.2 & 68.9  & 80.2   \\  \hline
		\multirow{2}{*}{RSPNet~\cite{chen2020rspnet}}    
		& C3D               & 36.0           & 56.7          & 66.5          & 76.3           & 87.7   \\
		& 3D R18         & 41.1  & 59.4 & 68.4 & 77.8  & 88.7 \\ 
		\hline
		Ours     & 3D R18            & \textbf{58.9}  & \textbf{76.3}  & \textbf{82.2} & \textbf{87.5}  & \textbf{93.4}  \\ 
		\shline
	\end{tabular}
	\caption{Comparison with state-of-the art methods for nearest neighbor retrieval task on the UCF-101 dataset as measured by the top-$k$ retrieval accuracy (\%).}
	\label{tab:sota-retri}
\end{table*}

\textbf{Different evaluation protocols.}
We survey existing self-supervised video representation learning methods and make the following observations about the evaluation protocols:
(1) Different works may use different cropping strategies for evaluation, such as center-crop~\cite{speednet,pace,chen2020rspnet}, three-crop~\cite{CVRL}, and ten-crop~\cite{DPC,han2020memdpc,han2020coclr}.
(2) Even with the same backbone, many methods may use different resolutions (\ie $112^2$, $128^2$, $224^2$, $256^2$) or numbers of frames (\ie 16, 32, 64) during evaluation.
For readers’ reference, in Table~\ref{t:test}, we present the results of our method with different evaluation protocols used in prior works.

\textbf{Impact of the pretraining epochs.}
We experiment with pretraining epochs varying from 100 to 400 and report the top-1 accuracy on UCF-101. Table~\ref{t:epoch} shows the impact of the number of training epochs on performance.
While ASC benefits from longer training, it already achieves strong performance after 200 epochs, \ie 80.52\%.
We also notice that performance starts to saturate after 300 epochs.

\textbf{Comparison with the state of the art.}
In Table~\ref{table:sota-cls}, we perform a thorough comparison with the state-of-the-art self-supervised learning methods and report the top-1 accuracy on UCF-101~\cite{ucf101} and HMDB-51~\cite{hmdb51}.
We show the pretraining settings for all approaches, \eg, pretraining dataset, backbone, number of input frames, resolution, and whether or not only the RGB modality is used.
Here, we mainly list the models using RGB as inputs for fair comparisons.
Since prior works use different backbones for experiments, we provide results of our ASCNet trained with two common architecture, \ie 3D ResNet-18~\cite{3dresnet}, S3D-G~\cite{s3dg}.

Our ASCNet achieves state-of-the-art results on both the UCF-101 and HMDB-51 datasets.
Specifically, when pretrained with the 3D ResNet-18 backbone, our method outperforms 3D-RotNet~\cite{3DRotNet}, ST-Puzzle~\cite{3dpuzzle}, and DPC~\cite{DPC} by a large margin (\textbf{80.5\%} \emph{vs.} 62.9\%, 65.8\%, and 68.2\%, respectively, on UCF-101 and \textbf{52.3\%} \emph{vs.} 33.7\%, 33.7\%, and 34.5\%).
When utilizing S3D-G as the backbone, our ASCNet achieves better accuracy than SpeedNet~\cite{speednet}, Pace~\cite{pace}, and RSPNet~\cite{chen2020rspnet} (\textbf{90.8\%} \emph{vs.} 81.1\%, 87.1\%, and 89.9\%, respectively, on UCF-101 and \textbf{60.5\%} \emph{vs.} 48.8\%, 52.6\%, and 59.9\%) under the same settings.
Remarkably, without the need of any annotation for pretraining, our ASCNet outperforms the ImageNet~\cite{deng2009imagenet} supervised pretrained model over two datasets (\textbf{90.8\%} \emph{vs.} 86.6\%, \textbf{60.5\%} \emph{vs.} 57.7\%).

\subsection{Evaluation on the Video Retrieval Task}
\begin{figure}
    \centering
    \includegraphics[width=0.46\textwidth]{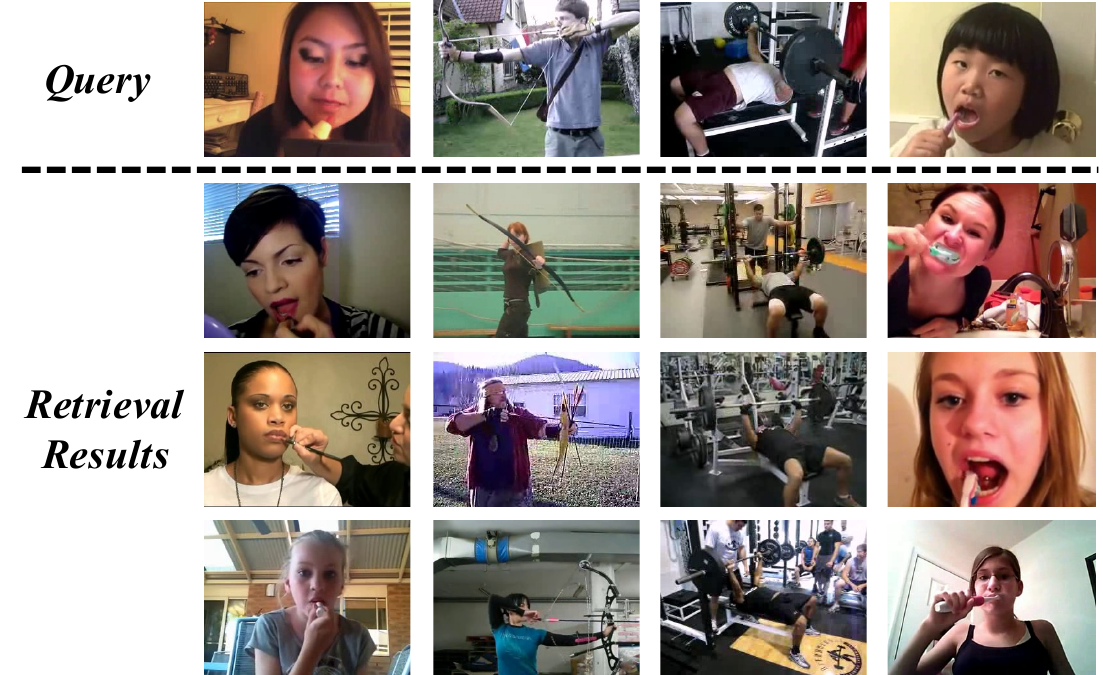}
    \caption{Qualitative examples of the video retrieval task.}
    \label{fig:retrieval}
\end{figure}
\textbf{Comparison with the state of the art.}
To further verify the effectiveness of ASCNet, we evaluate our representation with nearest neighbor video retrieval.
Specifically, following prior works~\cite{pace,speednet}, we uniformly sample 10 clips for each video. For all clips, features are extracted from the video encoder, which is only pretrained with self-supervised learning, and no further fine-tuning is allowed.
Then, we perform average-pooling over 10 clips to obtain a video-level feature vector.
We use each clip in the test set to query the \emph{k} nearest clips in the training set.
Experiments are conducted on the UCF-101 dataset and we evaluate our method on the split 1 of UCF-101 dataset and apply the top-\emph{k} accuracies (\emph{k} = 1, 5, 10, 20, 50) as the evaluation metrics.
As shown in Table~\ref{tab:sota-retri}, with the same 3D ResNet-18 backbone, our ASCNet outperforms the state-of-the-art method equivalent on all of the metrics by substantial margins (10.7\% - 45.9\% for top-1 accuracy on UCF-101).
These results indicate that the proposed pretext tasks help us to learn more discriminative features.

\textbf{Qualitative results for video retrieval.}
We further provide some retrieval results as a qualitative study.
In Figure~\ref{fig:retrieval}, the top is the query video from the UCF-101 testing set, and the bottom shows the top-3 nearest neighbors from the UCF-101 training set.
We successfully retrieve highly relevant videos with similar appearance and motion. This result implies that our method is able to learn both meaningful appearance and motion features for videos.

\section{Conclusion}

This work presents an unsupervised video representation learning framework named ASCNet, which leverages the appearance consistency throughout the frames of the same video and speed consistency between videos with the same fps. We train the model to map these clips to appearance and speed embedding space while maintaining consistency. We also propose an appearance-based retrieval strategy to reduce the conflict between the appearance and speed consistency perception tasks. Extensive experiments show that the features learned by ASCNet perform better on action recognition and video retrieval tasks. In the future, we plan to incorporate additional modalities into our framework.

\paragraph{Acknowledgments.} This work was partially supported by
National Natural Science Foundation of China (NSFC) 62072190,
Ministry of Science and Technology Foundation Project (2020AAA0106901),
Program for Guangdong Introducing Innovative and Entrepreneurial Teams 2017ZT07X183,
CCF-Baidu Open Fund (CCF-BAIDU OF2020022).

{\small
\bibliographystyle{ieee_fullname}
\bibliography{egbib}
}

\end{document}